\documentclass[letterpaper,10pt,conference]{ieeeconf}

\IEEEoverridecommandlockouts
\usepackage{graphicx}
\usepackage{amsmath}
\usepackage{amssymb}
\usepackage{booktabs}
\usepackage{array}
\usepackage{cite}
\usepackage{url}
\usepackage{xcolor}
\usepackage{flushend}
\usepackage{microtype}

\title{\LARGE \bf
DEXTERA: From a Single Image to Deployable Dexterous Manipulation via Real-to-Sim-to-Real
}

\author{Jin Wu$^{1}$, Lianjie Yuan$^{1,\dagger}$, Zeyan Sun$^{1,\dagger}$,
Yuanyuan Lei$^{2}$, Disi A$^{3}$, Bicheng Han$^{3}$, and Fangzhou Xia$^{1,*}$%
\thanks{$^{\dagger}$Lianjie Yuan and Zeyan Sun contributed equally.
$^{*}$Corresponding author.}%
\thanks{$^{1}$The University of Texas at Austin; $^{2}$University of Florida;
$^{3}$BrainCo.}%
\thanks{First-author email: \texttt{jinnwu@utexas.edu}. Corresponding-author
email: \texttt{fangzhou.xia@austin.utexas.edu}.}%
}

\begin{document}
% Compact author display; the control entry is not a numbered reference.
\bstctlcite{brief:control}

\maketitle
\thispagestyle{empty}
\pagestyle{empty}
\raggedbottom

\begin{abstract}
Collecting real-world robot data for dexterous manipulation is costly and time-consuming. While high-fidelity physics simulators enable scalable data synthesis and policy learning, constructing deployment-ready digital twins manually remains labor-intensive, and residual visual, geometric, and dynamics gaps hinder reliable sim-to-real transfer. We present DEXTERA, an automated real-to-sim-to-real framework that transforms a single RGB image into deployable policies for dexterous manipulation across four unified stages: (1) single-image scene factorization into a static Gaussian background and interactive rigid or articulated assets with VLM-inferred physical parameters; (2) metric scene global alignment, object canonicalization, and morphology-balanced robot calibration; (3) scalable simulator task primitive construction, VR teleoperation, and object-centric trajectory synthesis; and (4) a shared multimodal policy interface supporting both imitation learning and reinforcement learning. We evaluate DEXTERA across 13 task--embodiment pairs, 2 dexterous robot platforms, and 6 policy architectures. Experimental results demonstrate that DEXTERA achieves superior visual fidelity and 3D geometric reconstruction compared to generative baselines, while cross-domain trajectory replays validate strong physical interaction consistency. Furthermore, simulation-only trained policies enable viable zero-shot real-robot deployment, while simulation--real co-training substantially improves mean physical policy success from 29.2\% to 61.9\% across diverse policy architectures. Project website: \url{https://dextera-project.github.io/}.
\end{abstract}

\section{INTRODUCTION}

Large robot foundation models have broadened the range of manipulation behaviors learned across tasks and embodiments~\cite{kim2024openvla,black2025pi05}. Their progress, however, still depends on high-quality trajectories whose collection requires physical hardware, human operators, repeated resets, and careful safety supervision~\cite{tang2026modevla,cheng2025opentelevision,wang2024dexcap}. These collection costs are amplified in dexterous manipulation, where high-dimensional actions and contact-rich interactions demand precise demonstrations and repeated trials. Data collection and evaluation therefore remain major barriers to scalable robot learning.

Simulators~\cite{makoviychuk2021isaacgym,mittal2025isaaclab,zakka2026mjlab} offer photorealistic rendering and physics simulation, while simulation-based benchmarks provide reproducible platforms for scalable data collection, policy evaluation, and scene synthesis~\cite{james2020rlbench,liu2023libero,li2024behavior1k,wang2024gensim,chen2025robotwin2}. However, they generally assume that suitable assets and scenes already exist. Reproducing a deployment workspace still requires manual mesh authoring, physical-parameter specification, and camera-robot alignment to match the real world both visually and physically. Domain randomization and augmentation improve robustness, but cannot fully capture unmodeled appearance, sensing noise, and interaction dynamics, leaving a persistent sim-to-real gap~\cite{tobin2017domainrandomization,peng2018dynamicsrandomization,yuan2024maniwhere}.

Real-to-sim construction addresses this upstream bottleneck by converting observations into simulation-ready environments. Early approaches primarily reconstruct scene geometry and appearance from multi-view images or videos, using representations such as NeRFs and 3D Gaussian splats~\cite{mildenhall2020nerf,kerbl2023gaussians,torne2024rialto}. More recent generative methods reduce capture requirements by completing occluded geometry and reconstructing object-centric scenes from sparse or even single-view observations~\cite{sautter2025regen,yao2025cast,wang2026tabletopgen}. Beyond static reconstruction, emerging approaches incorporate physical properties to enable interactive, physics-aware scenes~\cite{chen2025physgen3d,zhang2026physomni,li2025wonderplay}.

However, an interactive reconstruction alone is insufficient for dexterous manipulation. A practical pipeline also needs accurate robot-scene alignment, physically valid interactions, task construction, and support for real-world deployment. Recent systems connect reconstruction with robot manipulation~\cite{qureshi2025splatsim,han2026re3sim,ranawaka2026simfoundry,wang2026tabletopgen}, but integration remains incomplete and physical evaluations predominantly concern parallel-jaw manipulation (Section~\ref{sec:related-r2s}). Extending these frameworks to dexterous hands, with more contact-rich dynamics and higher-dimensional control, remains underexplored.

This leaves a key question: can a single image be transformed into a visually faithful, physically grounded, and robot-aligned environment that supports multimodal policy learning and sim-to-real transfer for dexterous manipulation?

We present DEXTERA, a unified framework that turns a single workspace image into deployable policies for dexterous manipulation through a real-to-sim-to-real pipeline (Fig.~\ref{fig:asset-factorization}--\ref{fig:task-construction}). Given a single image, DEXTERA first reconstructs the workspace as a photorealistic background together with interactive rigid and articulated objects, including their visual geometry, collision models, and physical properties. It then automatically calibrates the reconstructed scene to the real workspace by aligning scene and object scales, camera pose, and robot geometry. Based on the reconstructed objects, reusable task primitives are used to create manipulation tasks automatically. Multimodal demonstrations are subsequently collected through VR teleoperation and object-centric trajectory synthesis, providing training data for both imitation learning (IL) and reinforcement learning (RL). Finally, the shared policy interface supports physical deployment on the real robot setup after training.

Specifically, this work makes the following contributions:
\begin{itemize}
    \item DEXTERA generates textured object meshes, collision geometry, and physical defaults from a single image and then aligns object scale, canonical coordinate frame, and camera pose with the real-world setup. Experiments further demonstrate superior visual fidelity compared with baselines and validate its physical consistency with an 87.95\% success rate in cross-domain replay on contact-rich tasks.
    \item DEXTERA introduces reusable task primitives that define scalable manipulation tasks from the reconstructed objects; it also provides a VR teleoperation interface, trajectory synthesis, and multimodal dataset collection pipelines that support diverse IL and RL benchmarks.
    \item Extensive real-to-sim-to-real experiments span 12 manipulation tasks, 2 robot configurations, and 6 IL/RL algorithms. Further experiments also show that mixing limited real-world data significantly improves policy performance over simulation-only training.
\end{itemize}

\section{RELATED WORK}

\subsection{3D Scene Synthesis}

Recent advances in neural rendering and image-conditioned generation enable high-fidelity 3D scene and object reconstruction from sparse views~\cite{mildenhall2020nerf,kerbl2023gaussians,liu2023zero123,xu2024instantmesh,zhao2025hunyuan3d20,yang2025hunyuan3d21,hyworld2026,worldlabs2025marble}. However, robotic manipulation demands representations beyond mere visual priors: objects must be actionable, support collision, and align metrically with the real world. While component-aware methods begin to address this by modeling scenes as interactive and editable instances rather than monolithic structures or collision-free layouts~\cite{sautter2025regen,yao2025cast,wang2026tabletopgen}, they typically fail to ground these environments in the physical workspace. Furthermore, physical interaction necessitates inferring articulation and dynamics directly from visual inputs~\cite{zhang2023partscene,liu2025singapo,guo2026articulat3d,li2026particulate,chen2025physgen3d,zhang2026physomni,li2025wonderplay}. Building on these insights, DEXTERA reconstructs simulation-ready environments from a single image. Decomposing the observation into static backgrounds and interactive rigid or articulated assets supports subsequent metric alignment with the physical robot workspace.

\subsection{Real-to-Sim for Robot Learning}
\label{sec:related-r2s}

Real-to-sim methods aim to bridge the deployment gap by constructing digital twins in simulation. While recent systems enhance visual fidelity and robustify policies through reconstructed geometry, they often rely on multi-view scans, additional reconstruction procedures, or task-specific pipelines~\cite{qureshi2025splatsim,torne2024rialto,han2026re3sim}. Recent frameworks move toward integrated workflows but face constraints. For instance, SimFoundry requires multi-view video capture~\cite{ranawaka2026simfoundry}, while TabletopGen and RoboSnap focus their setup on tabletop manipulation with parallel-jaw grippers~\cite{wang2026tabletopgen,zhang2026robosnap}. Extending such workflows to single-image capture, structured and scalable task generation, and dexterous, contact-rich manipulation therefore remains an open challenge. In contrast, DEXTERA generates assets from a single
image, introduces reusable task primitives that scale to manipulation
benchmarks, and supports dexterous hand embodiments.

\subsection{Simulation-Based Task and Policy Scaling}

Modern simulation frameworks and benchmarks have standardized task definitions, data generation, and policy evaluation~\cite{coumans2016pybullet,todorov2012mujoco,xiang2020sapien,makoviychuk2021isaacgym,mittal2025isaaclab,james2020rlbench,liu2023libero,li2024behavior1k}. While automated task synthesis, trajectory adaptation, and domain randomization have significantly scaled data collection and robustness~\cite{wang2024gensim,chen2025robotwin2,mandlekar2023mimicgen,tobin2017domainrandomization,peng2018dynamicsrandomization,yuan2024maniwhere}, these approaches predominantly operate on manually predefined assets rather than environments reconstructed from a specific physical workspace. Consequently, scene construction, task generation, and policy learning~\cite{pinto2018asymmetric,mandlekar2021robomimic,chi2023diffusionpolicy,kim2024openvla} are typically treated as decoupled stages. DEXTERA bridges this gap, unifying automatic scene and asset reconstruction with scalable task generation and multimodal data collection to provide a seamless workflow from single-image observation to real-world policy deployment.

\begin{figure}[!t]
\centering
\includegraphics[width=\columnwidth]{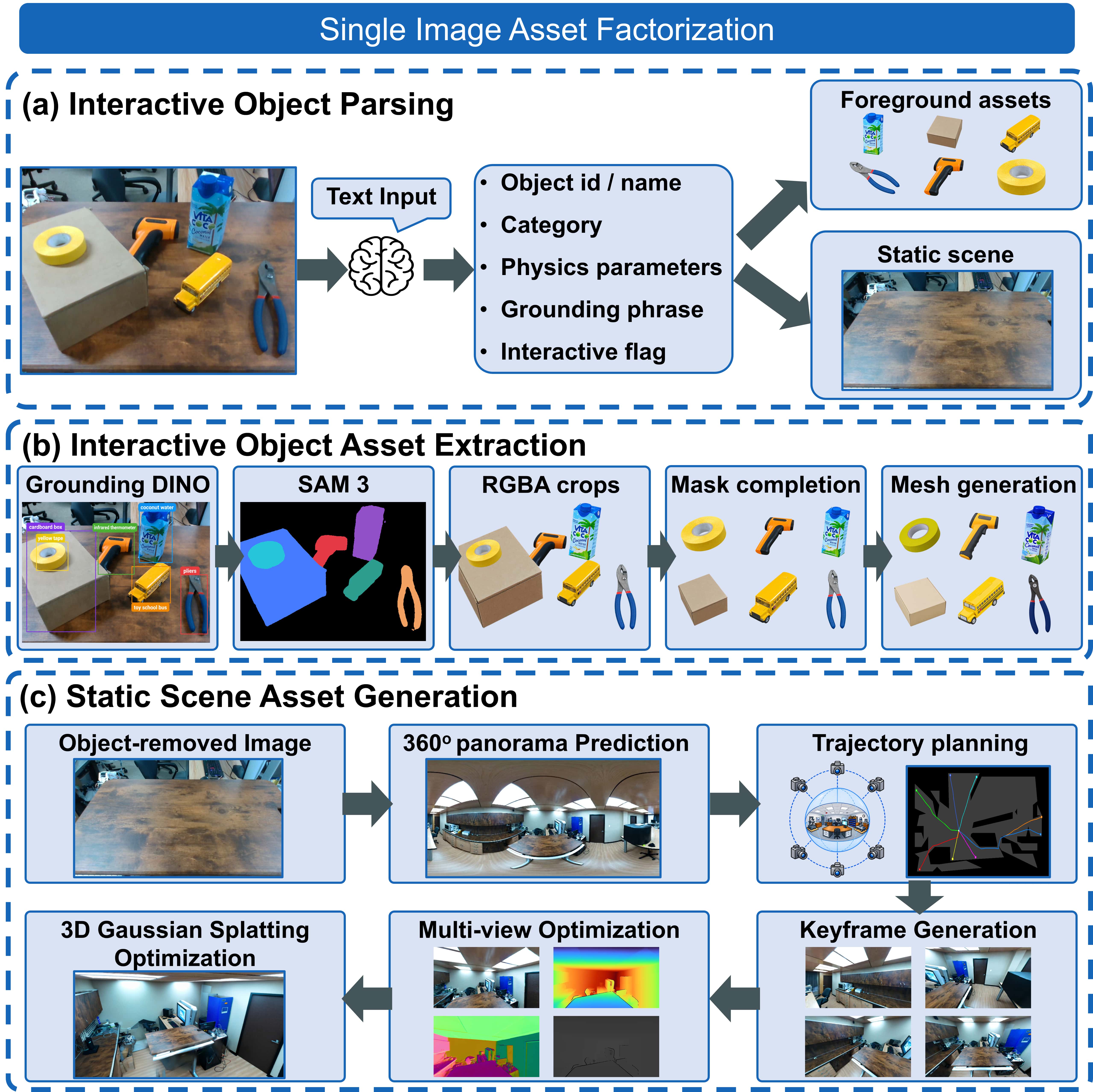}
\caption{Single-image asset factorization. (a) Semantic parsing separates interactive objects from the static scene. (b) Object localization, segmentation, completion, and mesh generation produce foreground assets. (c) Background completion and multi-view reconstruction produce the Gaussian scene.}
\label{fig:asset-factorization}
\end{figure}

\section{Method}

\begin{figure}[!t]
\centering
\includegraphics[width=\columnwidth]{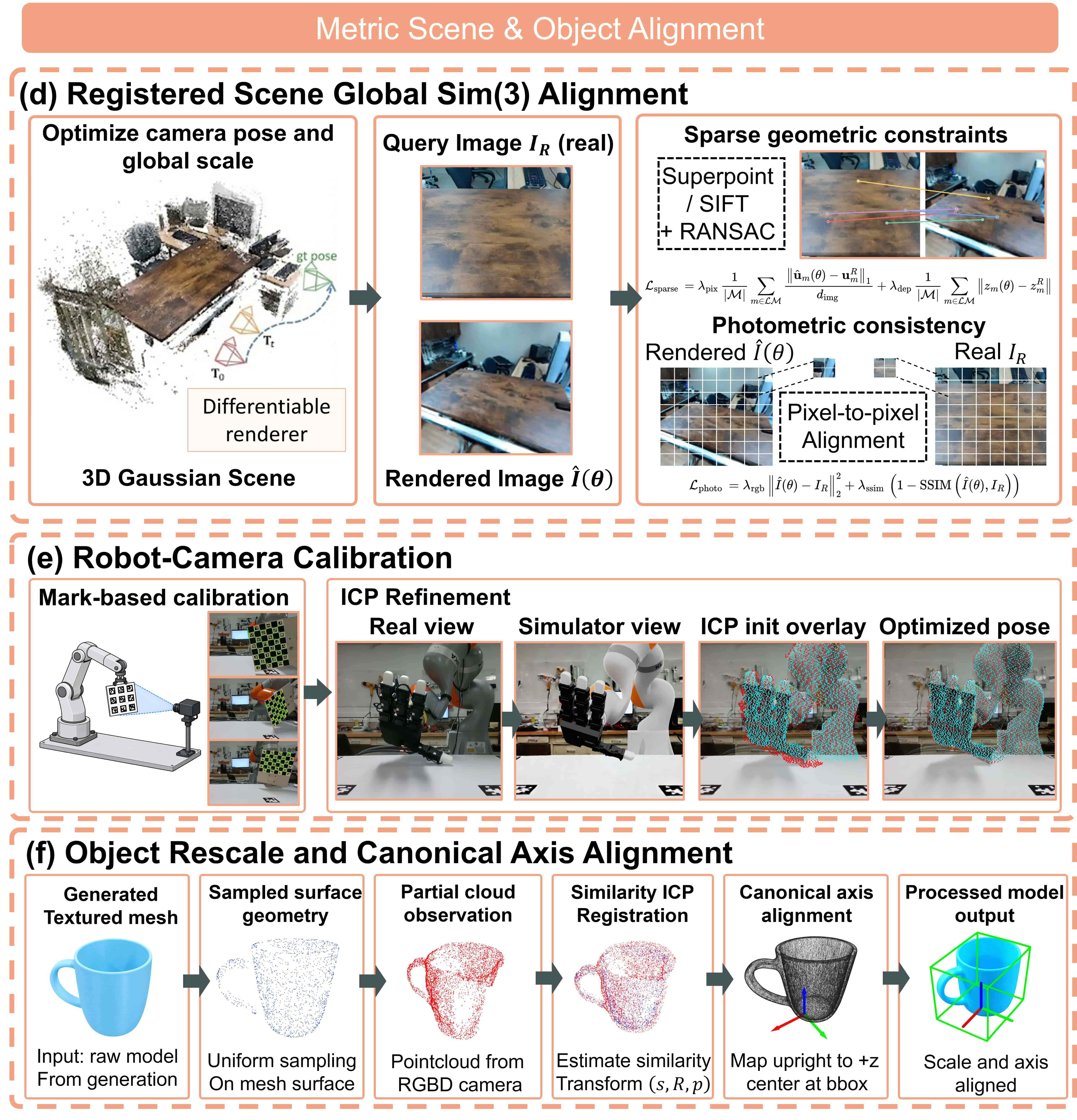}
\caption{Metric scene and object alignment. (d) Photometric and sparse geometric constraints recover scene scale and camera pose. (e) Robot geometry refines marker-based calibration. (f) Object rescaling and canonical-axis alignment ground each asset metrically.}
\label{fig:metric-alignment}
\end{figure}

DEXTERA receives a single RGB scene image $I_0$, a robot model, and a calibration capture. The capture contains camera intrinsics, synchronized robot poses, and depth observations. Only $I_0$ generates the assets, while the additional measurements are used to recover metric scale and align them with the robot setup. These stages are detailed in Fig.~\ref{fig:asset-factorization}--\ref{fig:task-construction}.

\subsection{Single-Image Asset Factorization}

As shown in Fig.~\ref{fig:asset-factorization}, we first decompose the input image into two complementary representations: independent interactive object assets and a static scene background, each producing a distinct 3D representation required for downstream simulation.

\paragraph{Interactive Object Parsing}
In Fig.~\ref{fig:asset-factorization}(a), we prompt Qwen3-VL~\cite{bai2025qwen3vl} to parse $I_0$ into scene elements. For each scene element, it predicts a name, category, grounding phrase, body type, physical defaults, and interactive flag. Interactive elements are routed to the foreground assets branch, while support surfaces, room structure, and other fixed elements remain in the static background set.

\paragraph{Interactive Object Asset Extraction}
During the stage illustrated in Fig.~\ref{fig:asset-factorization}(b), Grounding DINO first localizes each interactive object from its grounding phrase~\cite{liu2024groundingdino}. Duplicate detections are suppressed when multiple phrases refer to the same image region. SAM 3 then extracts an instance mask from the selected box and object name~\cite{carion2026sam3}. For heavily occluded masks, FLUX.2 completes the missing image content conditioned on the object name and visible mask features~\cite{blackforestlabs2025flux2}. Finally, the image-to-3D model, Hunyuan3D 3.1, converts each object mask into an independent textured mesh~\cite{tencentcloud2026hunyuan3d31}, while a separate configuration file stores the object's predicted properties.

The predicted body type determines the final asset representation. Rigid objects are kept as single-body meshes, while articulated objects are further processed by a feed-forward model to infer link segmentation, kinematic structure, joint types, axes, and motion limits~\cite{li2026particulate}. These predictions are associated with the corresponding link meshes and exported as articulated URDF assets.

\paragraph{Static Scene Asset Generation}
For the static branch in Fig.~\ref{fig:asset-factorization}(c), we remove all interactive objects and the robot from $I_0$ using FLUX.2~\cite{blackforestlabs2025flux2}, retaining only the static background context. HY-World 2.0 then expands the cleaned image into a 360$^\circ$ panorama, plans virtual camera trajectories, and synthesizes view-consistent observations along these trajectories~\cite{hyworld2026}. The generated multi-view RGB images, together with predicted depth, normals, and masks, are used to optimize a 3D Gaussian representation of the background. The resulting 3DGS provides photorealistic appearance, while the associated point cloud and mesh provide explicit geometry for collision modeling and subsequent physics interactions. The background and foreground assets form the library for subsequent alignment.

\subsection{Metric Scene and Object Alignment}
While visually consistent with $I_0$, the generated assets initially reside in arbitrary coordinate frames. To tie them to the physical workspace, we perform a tri-level alignment procedure that grounds the static scene, robot, and interactive objects into a unified metric world frame $W$ (Fig.~\ref{fig:metric-alignment}).

\paragraph{Static scene alignment}
We align the reconstructed Gaussian background to a measured RGB-D observation by optimizing its scale $s_G$ and camera pose ${}^{G}\mathbf T_C$. Let $\theta=({}^{G}\mathbf T_C,s_G)$. As illustrated in Fig.~\ref{fig:metric-alignment}(d), the optimization minimizes a hybrid objective merging dense photometric alignment with sparse geometric constraints:
\begin{equation}
\begin{aligned}
\mathcal L_{\mathrm{scene}}(\theta)
&=\mathcal L_{\mathrm{photo}}(\theta)
  +\mathcal L_{\mathrm{sparse}}(\theta),\\
\mathcal L_{\mathrm{photo}}(\theta)
&=\lambda_{\mathrm{rgb}}
  \|\hat I(\theta)-I_R\|_{\Omega}^{2}\\
&\quad+\lambda_{\mathrm{ssim}}
  \bigl[1-\mathrm{SSIM}(\hat I(\theta),I_R)\bigr],\\
\mathcal L_{\mathrm{sparse}}(\theta)
&=\frac{\lambda_{\mathrm{pix}}}{|\mathcal M|}
  \sum_{m\in\mathcal M}
  \frac{\|\hat{\mathbf u}_m(\theta)-\mathbf u_m^R\|_1}
       {d_{\mathrm{img}}}\\
&\quad+\frac{\lambda_{\mathrm{dep}}}{|\mathcal M|}
  \sum_{m\in\mathcal M}
  |\hat z_m(\theta)-z_m^R|.
\end{aligned}
\label{eq:scene-calibration}
\end{equation}

% Here, $\mathcal L$ denotes loss for different terms; $\lambda_{\mathrm{rgb}}$, $\lambda_{\mathrm{ssim}}$, $\lambda_{\mathrm{pix}}$, and $\lambda_{\mathrm{dep}}$ are the corrsponding weights. The photometric term compares rendered $\hat I(\theta)$ and measured $I_R$ over valid pixels $\Omega$ using RGB and structural similarity (SSIM). The sparse term uses RANSAC-filtered SuperPoint correspondences $\mathcal M$~\cite{detone2018superpoint}: rendered and measured pixel locations $\hat{\mathbf u}_m(\theta)$ and $\mathbf u_m^R$ constrain camera pose, with errors normalized by image diagonal $d_{\mathrm{img}}$; their depths $\hat z_m(\theta)$ and $z_m^R$ constrain metric scale. Valid-depth filtering retains measured geometry. Applying the optimized transform and scale to both Gaussians and the collision mesh preserves alignment between appearance and simulated contacts.

Here, $\mathcal L$ denotes loss for different terms and the $\lambda$ coefficients are the corresponding weights. The photometric objective $\mathcal L_{\mathrm{photo}}$ enforces appearance consistency between the rendered image $\hat I(\theta)$ and the measured observation $I_R$ over the valid pixel domain $\Omega$ using RGB color differences and structural similarity (SSIM). Concurrently, the sparse geometric objective $\mathcal L_{\mathrm{sparse}}$ leverages RANSAC-filtered SuperPoint correspondences $\mathcal M$~\cite{detone2018superpoint} to anchor spatial metrics. The normalized pixel reprojection residual between rendered locations $\hat{\mathbf u}_m(\theta)$ and measured coordinates $\mathbf u_m^R$ constrains the camera pose, while the depth error between $\hat z_m(\theta)$ and valid measured depth $z_m^R$ explicitly bounds the metric scene scale. Finally, the optimized scale $s_G$ and transformation ${}^{G}\mathbf T_C$ are applied to both the 3D Gaussian representations and collision meshes.

\paragraph{Robot-Camera Calibration}
As shown in Fig.~\ref{fig:metric-alignment}(e), we refine the calibration from a coarse marker-based initialization, ${}^{C}\mathbf T_B^{(0)}$~\cite{garrido2014aruco,tsai1989handeye}, directly against the observed point cloud. Rendered robot masks acquired  from Isaac Lab and real robot masks extracted by SAM 3~\cite{carion2026sam3} are back-projected into 3D for matching. To prevent alignment bias toward the spatially dominant arm geometry and ensure hand calibration accuracy, we sample an equal number of points $N$ from the arm and hand to perform morphology-balanced ICP~\cite{besl1992icp}:
\begin{equation}
\begin{aligned}
\mathcal L_c(\Delta\mathbf T)
&=\frac{1}{N}\sum_{j=1}^{N}
\|\Delta\mathbf T\mathbf p_{c,j}^{S}
-\mathbf p_{c,j}^{R}\|_2^2,\\
\Delta\mathbf T^*
&=\arg\min_{\Delta\mathbf T}
\frac{\mathcal L_{\mathrm{arm}}+\mathcal L_{\mathrm{hand}}}{2}.
\end{aligned}
\label{eq:balanced-robot-icp}
\end{equation}
Here, $c\in\{\mathrm{arm},\mathrm{hand}\}$ indexes robot components; $\mathbf p_{c,j}^{S}$ and $\mathbf p_{c,j}^{R}$ are corresponding simulated and real 3D points in camera coordinates.  The rigid transformation $\Delta\mathbf T \in \mathrm{SE}(3)$ optimizes the relative offset by minimizing the component-balanced loss, yielding the optimal correction $\Delta\mathbf T^*$.  Finally, with $B$ denoting the robot base frame, the refined camera-to-base extrinsic matrix ${}^{C}\mathbf T_B^{*}$ is updated from the initial estimate:
\begin{equation}
{}^{C}\mathbf T_B^{*}
=\Delta\mathbf T^{*}{}^{C}\mathbf T_B^{(0)}.
\label{eq:refined-camera-robot}
\end{equation}

\paragraph{Object Rescaling, Canonicalization, and Tracking}
As detailed in Fig.~\ref{fig:metric-alignment}(f), interactive assets recover metric scale by computing bounding-extent ratios between back-projected RGB-D observations and reconstructed meshes, followed by Sim(3) ICP registration. For articulated assets, this scale factor is uniformly propagated across all link geometries and joint pivots to preserve kinematic consistency. To establish a standardized object-local coordinate frame, which is essential for scalable task generation in later stages, Qwen3-VL aligns candidate bounding-box axes against reference image crops to resolve semantic orientations (e.g., mapping the functional upright axis to $+Z$). This canonical transformation is likewise applied across all kinematic sub-components. Finally, objects are continuously tracked in $W$ via FoundationPose~\cite{wen2024foundationpose}, initialized by Grounding DINO and SAM 3, with DeAOT~\cite{yang2022deaot} suppressing temporal tracking drift.

\begin{table}[!t]
\caption{Observation modalities exposed by the task interface.}
\label{tab:observation-modalities}
\centering
\scriptsize
\setlength{\tabcolsep}{3pt}
\begin{tabular}{@{}p{0.22\columnwidth} p{0.70\columnwidth}@{}}
\toprule
Modality & Components \\
\midrule
Policy & Previous robot action. \\
Proprioception & Robot joint positions. \\
State & Object poses and orientation, articulated-joint displacement. \\
Contact & 3D contact force at each fingertip. \\
Goal & Target pose and orientation, signed articulated-joint goal position. \\
Privileged & Robot joint and object velocities, hand-base and fingertip poses and spatial velocities, sampled object surface points. \\
RGB & RGB images from the available cameras. \\
Depth & Image-plane distance maps from the available cameras. \\
Point cloud & Camera-frame XYZ points back-projected from depth. \\
\bottomrule
\end{tabular}
\end{table}

\subsection{Simulator Task Construction}
\begin{figure}[!t]
\centering
\includegraphics[width=\columnwidth]{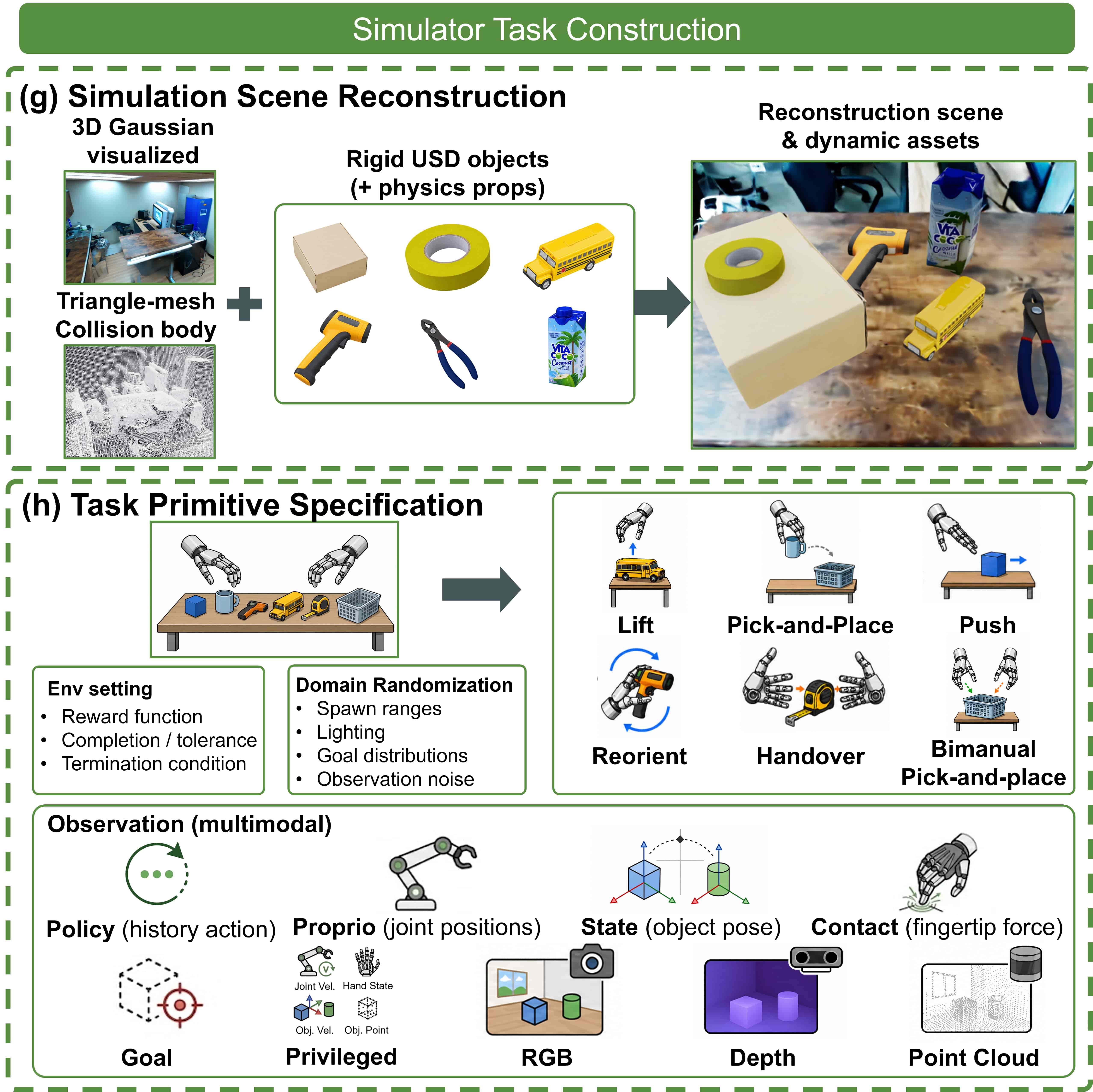}
\caption{Simulator task construction. (g) Gaussian appearance, mesh collisions, and dynamic assets form the simulation scene. (h) Task primitives define objectives, randomization, and success criteria, with shared multimodal observations for learning and deployment.}
\label{fig:task-construction}
\end{figure}

In this stage, aligned robots and objects are imported into Isaac Lab as USD assets~\cite{mittal2025isaaclab} as visualized in Fig.~\ref{fig:task-construction}(g). The background supplies Gaussian appearance and static mesh collisions. Predicted mass, friction, restitution, and stiffness are assigned to the corresponding assets during initialization. Each task combines an embodiment, reconstructed assets, and one of six primitives (Fig.~\ref{fig:task-construction}(h)). The primitive defines reset distributions, reward functions, and success criteria as below, while the embodiment specifies active joints and robot type.

\begingroup\footnotesize\raggedright
\begin{list}{$\bullet$}{\setlength{\leftmargin}{1.2em}\setlength{\itemindent}{0pt}\setlength{\labelwidth}{0.7em}\setlength{\labelsep}{0.5em}\setlength{\itemsep}{3pt}\setlength{\parsep}{0pt}\setlength{\topsep}{3pt}\setlength{\partopsep}{0pt}}
\item \textbf{Lift:}  Object exceeds target height and maintains a valid grasp.
\par\emph{Examples:} \texttt{LiftToyBus}, \texttt{LiftBlueMug}, \texttt{BimanualLiftBasket}.
\item \textbf{Push:}  Object reaches target position and orientation within tolerance.
\par\emph{Examples:} \texttt{PushBlueCube}, \texttt{PushT}.
\item \textbf{Pick and reorient:}  Object reaches target position and orientation within tolerance, maintaining a valid grasp.
\par\emph{Examples:} \texttt{ReorientThermometer}, \texttt{ReorientCamera}.
\item \textbf{Pick and place:}  Object reaches target pose within tolerance or lies fully within the target container.
\par\emph{Examples:} \texttt{PlaceToyBusInBasket}, \texttt{StackBowls}.
\item \textbf{Handover:}  Receiving hand establishes and maintains a valid grasp after the source hand releases the object.
\par\emph{Example:} \texttt{HandoverGrayBowl}.
\item \textbf{Articulation:}  All selected joints reach commanded states within the prescribed joint-space tolerance.
\par\emph{Examples:} \texttt{CloseLaptop}, \texttt{OpenScissors}, \texttt{OpenSunglasses}.
\end{list}
\endgroup

To enhance policy robustness, domain randomization is applied to the object and goal poses, physical dynamics, robot configurations, and visual conditions (e.g., camera extrinsics, lighting, and noise). Table~\ref{tab:observation-modalities} summarizes the exposed observation spaces, from which downstream algorithms can select the inputs they require. 
Crucially, task definitions remain identical across simulation and hardware to ensure consistent evaluation. Furthermore, the observation interface strictly decouples deployable sensor measurements from privileged simulation states, seamlessly supporting both privileged teacher training in simulation and sensor-conditioned policy deployment on physical hardware.

\begin{table*}[!t]
\caption{Object-instance reconstruction and alignment over 86 valid instances.}
\label{tab:object-reconstruction}
\centering
\scriptsize
\setlength{\tabcolsep}{3.2pt}
\begin{tabular*}{\textwidth}{@{\extracolsep{\fill}}lcccccccc@{}}
\toprule
& \multicolumn{2}{c}{Segmentation and Localization} & \multicolumn{2}{c}{RGB Appearance} & \multicolumn{1}{c}{Depth} & \multicolumn{3}{c}{3D Reconstruction} \\
\cmidrule(lr){2-3}\cmidrule(lr){4-5}\cmidrule(lr){6-6}\cmidrule(lr){7-9}
Method & Mask IoU $\uparrow$ & Boundary F1 $\uparrow$ & RGB MAE $\downarrow$ & $\Delta E_{76}$ $\downarrow$ & \shortstack{Depth MAE\\(m) $\downarrow$} & \shortstack{Centroid\\(m) $\downarrow$} & \shortstack{Chamfer-L1\\(m) $\downarrow$} & F-score@1\,cm $\uparrow$ \\
\midrule
Ours & $\boldsymbol{.731\!\pm\!.136}$ & $\boldsymbol{.260\!\pm\!.120}$ & $\boldsymbol{.182\!\pm\!.062}$ & $\boldsymbol{27.90\!\pm\!7.64}$ & $\boldsymbol{.0093\!\pm\!.0099}$ & $\boldsymbol{.0115\!\pm\!.0160}$ & $\boldsymbol{.0069\!\pm\!.0076}$ & $\boldsymbol{.906\!\pm\!.132}$ \\
Hunyuan3D-2.1 & $.661\!\pm\!.180$ & $.239\!\pm\!.137$ & $.223\!\pm\!.063$ & $34.74\!\pm\!10.12$ & $.0130\!\pm\!.0115$ & $.0143\!\pm\!.0173$ & $.0103\!\pm\!.0122$ & $.750\!\pm\!.250$ \\
TRELLIS.2 & $.559\!\pm\!.196$ & $.163\!\pm\!.087$ & $.250\!\pm\!.063$ & $39.11\!\pm\!11.23$ & $.0296\!\pm\!.0179$ & $.0258\!\pm\!.0144$ & $.0199\!\pm\!.0120$ & $.346\!\pm\!.202$ \\
SAM3D+FP & $.604\!\pm\!.143$ & $.236\!\pm\!.118$ & $.407\!\pm\!.118$ & $58.68\!\pm\!13.02$ & $.0166\!\pm\!.0140$ & $.0183\!\pm\!.0173$ & $.0118\!\pm\!.0081$ & $.622\!\pm\!.235$ \\
SAM3D-only & $.488\!\pm\!.285$ & $.165\!\pm\!.114$ & $.391\!\pm\!.111$ & $55.95\!\pm\!12.95$ & $.0235\!\pm\!.0235$ & $.0299\!\pm\!.0358$ & $.0231\!\pm\!.0252$ & $.538\!\pm\!.339$ \\
\bottomrule
\end{tabular*}
\end{table*}

\begin{table*}[!t]
\caption{Static-background fidelity of the HY-World reconstruction from the calibrated top-camera view.}
\label{tab:background-reconstruction}
\centering
\scriptsize
\setlength{\tabcolsep}{3.2pt}
\begin{tabular*}{\textwidth}{@{\extracolsep{\fill}}ccccccccccc@{}}
\toprule
\multicolumn{4}{c}{RGB Appearance} & \multicolumn{4}{c}{Depth} & \multicolumn{3}{c}{3D Reconstruction} \\
\cmidrule(lr){1-4}\cmidrule(lr){5-8}\cmidrule(lr){9-11}
RGB MAE $\downarrow$ & \shortstack{PSNR\\(dB) $\uparrow$} & SSIM $\uparrow$ & $\Delta E_{76}$ $\downarrow$ & \shortstack{Depth MAE\\(m) $\downarrow$} & \shortstack{Depth RMSE\\(m) $\downarrow$} & AbsRel $\downarrow$ & $\delta_1$ $\uparrow$ & \shortstack{Centroid\\(m) $\downarrow$} & \shortstack{Chamfer-L1\\(m) $\downarrow$} & F-score@1\,cm $\uparrow$ \\
\midrule
0.0583 & 20.74 & 0.761 & 8.22 & 0.00516 & 0.00655 & 0.010 & 1.000 & 0.00100 & 0.00633 & 0.900 \\
\bottomrule
\end{tabular*}
\end{table*}

\begin{table}[!t]
\caption{Real replay of 30 simulation-successful rollouts per task across two robot embodiments.}
\label{tab:physical-replay}
\centering
\scriptsize
\setlength{\tabcolsep}{3pt}
\begin{tabular}{@{}p{0.62\columnwidth}cc@{}}
\toprule
Task & Real succ. & Rate (\%) \\
\midrule
OB: \texttt{LiftToyBus} & 25/30 & 83.33 \\
OB: \texttt{LiftBlueMug} & 30/30 & 100.00 \\
OB: \texttt{PlaceToyBusInBasket} & 28/30 & 93.33 \\
OB: \texttt{HandoverGrayBowl} & 24/30 & 80.00 \\
OB: \texttt{BimanualLiftBasket} & 25/30 & 83.33 \\
OB: \texttt{ReorientThermometer} & 22/30 & 73.33 \\
OB: \texttt{BimanualLiftBox} & 30/30 & 100.00 \\
OB: \texttt{PushBlueCube} & 24/30 & 80.00 \\
OB: \texttt{BimanualPlaceBoxInBasket} & 22/30 & 73.33 \\
OB: \texttt{CloseLaptop} & 30/30 & 100.00 \\
\midrule
KL: \texttt{PlaceCoconutWaterInBasket} & 26/30 & 86.67 \\
KL: \texttt{LiftMustardBottle} & 27/30 & 90.00 \\
KL: \texttt{CloseLaptop} & 30/30 & 100.00 \\
\midrule
Overall & 343/390 & 87.95 \\
\bottomrule
\end{tabular}
\par\vspace{2pt}
\noindent\parbox[t]{0.98\columnwidth}{\scriptsize OB: OpenArm + BrainCo Revo1; KL: KUKA + LEAP Hand.}
\end{table}

\begin{table*}[!t]
\caption{Policy success rates (\%) across 13 task--embodiment pairs. Each entry is simulation/real, evaluated over 100/20 trials. Imitation policies use simulation and real data; T--S is PPO teacher--student distillation.}
\label{tab:policy-benchmark}
\centering
\scriptsize
\setlength{\tabcolsep}{5pt}
\begin{tabular*}{\textwidth}{@{\extracolsep{\fill}}lcccccc@{}}
\toprule
& \multicolumn{5}{c}{Imitation learning} & Reinforcement learning \\
\cmidrule(lr){2-6}\cmidrule(lr){7-7}
Task pair & ACT & DP & BC-RNN & OpenVLA & $\pi_{0.5}$ & T--S \\
\midrule
OB: \texttt{LiftToyBus} & 67/55 & 71/50 & 5/0 & 14/5 & 73/65 & 53/20 \\
OB: \texttt{LiftBlueMug} & 82/100 & 89/90 & 19/5 & 35/10 & 85/100 & 21/10 \\
OB: \texttt{PlaceToyBusInBasket} & 24/15 & 32/0 & 6/0 & 42/10 & 36/10 & 16/0 \\
OB: \texttt{HandoverGrayBowl} & 75/65 & 15/0 & 0/0 & 17/0 & 63/40 & 53/10 \\
OB: \texttt{BimanualLiftBasket} & 70/80 & 69/45 & 27/5 & 8/0 & 84/90 & 62/30 \\
OB: \texttt{ReorientThermometer} & 30/55 & 20/15 & 3/0 & 8/0 & 46/50 & 3/0 \\
OB: \texttt{BimanualLiftBox} & 47/35 & 82/50 & 13/0 & 12/5 & 84/60 & 49/0 \\
OB: \texttt{PushBlueCube} & 41/50 & 15/35 & 32/10 & 19/20 & 33/55 & 38/10 \\
OB: \texttt{BimanualPlaceBoxInBasket} & 36/10 & 23/5 & 3/0 & 1/0 & 60/20 & 7/0 \\
OB: \texttt{CloseLaptop} & 46/80 & 87/100 & 27/45 & 50/45 & 99/100 & 96/30 \\
\midrule
KL: \texttt{PlaceCoconutWaterInBasket} & 52/25 & 3/0 & 23/0 & 10/0 & 10/5 & 57/15 \\
KL: \texttt{LiftMustardBottle} & 88/35 & 32/20 & 38/0 & 70/20 & 78/40 & 41/25 \\
KL: \texttt{CloseLaptop} & 100/100 & 57/75 & 90/80 & 83/65 & 91/100 & 82/75 \\
\midrule
Mean & 58.3/54.2 & 45.8/37.3 & 22.0/11.2 & 28.4/13.8 & 64.8/56.5 & 44.5/17.3 \\
\bottomrule
\end{tabular*}
\par\vspace{2pt}
\noindent\parbox[t]{\textwidth}{\scriptsize OB: OpenArm + BrainCo Revo1; KL: KUKA + LEAP Hand. Means weight task--embodiment pairs equally.}
\end{table*}

\subsection{Policy Learning and Deployment}
\looseness=-1
As depicted in Fig.~\ref{fig:task-construction}(h), the task and observation interface facilitates multimodal policy learning and real-world deployment. Human demonstrations are collected via VR teleoperation with Meta Quest 3, mapping wrist poses through inverse kinematics and finger motions via retargeting~\cite{qin2023anyteleop}. To scale dataset diversity without physical overhead, an initial set of ten demonstrations per task is expanded across randomized object and goal configurations using object-centric trajectory synthesis. Object frames are queried directly in the simulator, and estimated online via  object tracking in physical environments during replay.

\looseness=-1
Let ${}^{W}\mathbf T_{EE}^{\mathrm{src}}(t)$ be the source end-effector trajectory. The source and new task frames are ${}^{W}\mathbf T_F^{\mathrm{src}}$ and ${}^{W}\mathbf T_F^{\mathrm{new}}$. Here, $EE$, $F$, and $t$ denote the end effector, task frame, and time index, respectively. Waypoint positions $\mathbf p_i$ and unit-quaternion orientations $\mathbf q_i$ are interpolated with coefficient $\lambda\in[0,1]$ using linear blending and spherical linear interpolation ($\mathrm{SLERP}$). Zero-mean Gaussian noise $\boldsymbol\epsilon_t \sim \mathcal N(\mathbf 0,\boldsymbol\Sigma)$ is added to the transformed action $\mathbf a_t^{\mathrm{transformed}}$, while invalid synthesized trajectories are filtered out by collision checks and success criteria. We preserve the demonstrated pose relative to the task frame, interpolate waypoints, and inject action noise:
\begin{equation}
\begin{aligned}
{}^{W}\mathbf T_{EE}^{\mathrm{new}}(t)
&={}^{W}\mathbf T_F^{\mathrm{new}}
  ({}^{W}\mathbf T_F^{\mathrm{src}})^{-1}
  {}^{W}\mathbf T_{EE}^{\mathrm{src}}(t),\\
\mathbf p(\lambda)&=(1-\lambda)\mathbf p_i
  +\lambda\mathbf p_{i+1},\\
\mathbf q(\lambda)&=\mathrm{SLERP}
  (\mathbf q_i,\mathbf q_{i+1},\lambda),\\
\mathbf a'_t&=\mathbf a_t^{\mathrm{transformed}}
  +\boldsymbol\epsilon_t,
\quad \boldsymbol\epsilon_t\sim\mathcal N(\mathbf 0,\boldsymbol\Sigma),
\end{aligned}
\label{eq:trajectory-synthesis}
\end{equation}
The resulting dataset supports training across multiple learning architectures as detailed in Section~\ref{sec:experiments}.

For reinforcement learning, a PPO teacher trained on privileged simulation states and contact forces supervises a student policy restricted to deployable observations (e.g., RGB images and proprioception) under domain randomization~\cite{schulman2017ppo,pinto2018asymmetric}. During physical deployment, ROS 2 synchronizes real-time camera streams, robot states, and control commands while maintaining identical observation representations and action spaces to ensure seamless sim-to-real transfer.

\begin{figure}[!t]
\centering
\includegraphics[width=0.85\columnwidth]{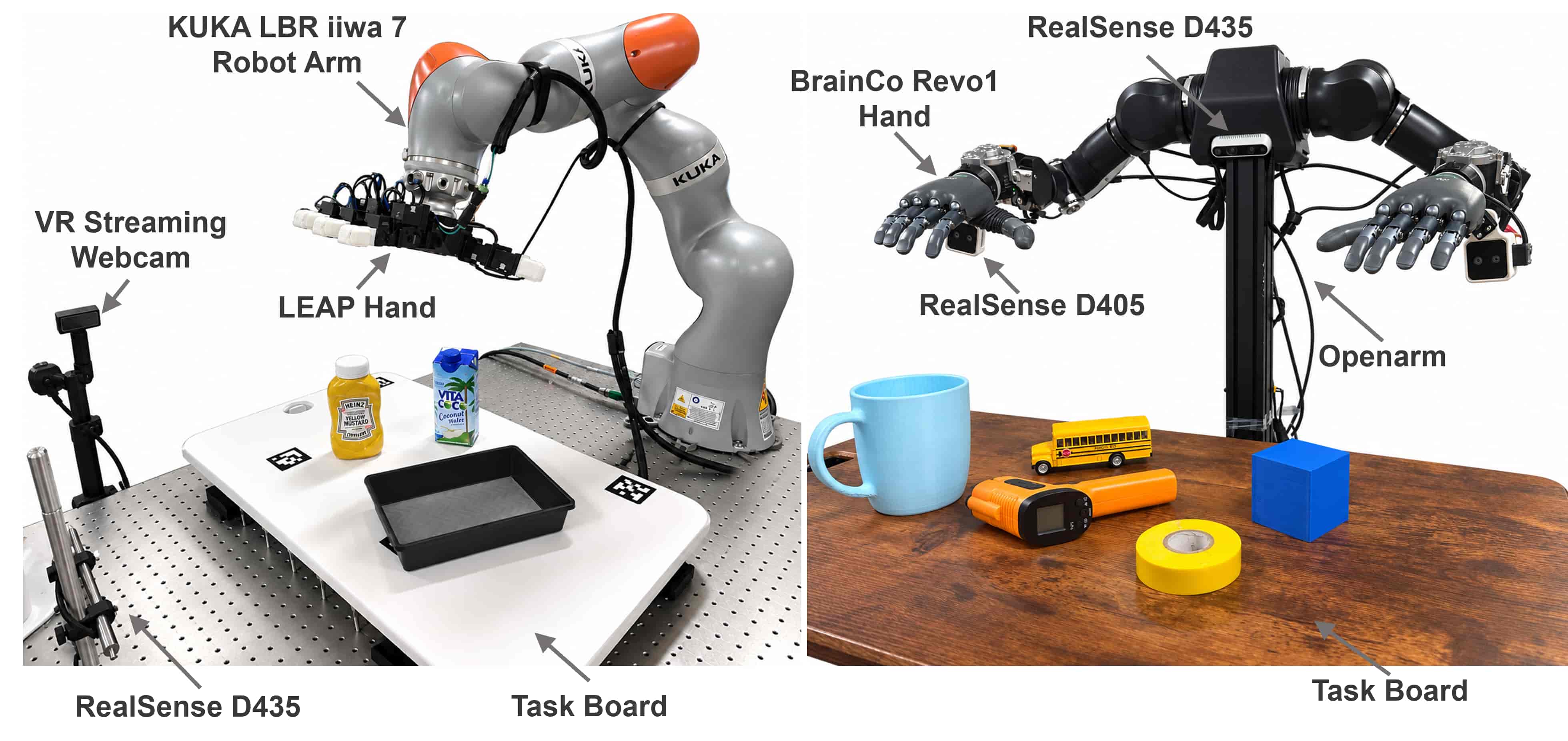}
\caption{Experimental platforms. Left: KUKA LBR iiwa 7 with a LEAP Hand. Right: dual-arm OpenArm with BrainCo Revo1 hands. The cameras and task workspaces are shown for each platform.}
\label{fig:experimental-setup}
\end{figure}

\section{Experiments}
\label{sec:experiments}
We evaluate DEXTERA across three main axes: (1) visual fidelity and 3D pose alignment against generative baselines; (2) interaction physical consistency via cross-domain trajectory replay; and (3) closed-loop policy learning and sim-to-real transfer. Fig.~\ref{fig:experimental-setup} shows the physical platforms: a single-arm KUKA LBR iiwa 7 with a LEAP Hand (KL) and a bimanual OpenArm with BrainCo Revo1 hands (OB).

\subsection{Visual Fidelity and Pose Alignment}
\label{sec:exp-reconstruction}
We evaluate object reconstruction on 18 RGB-D snapshots containing 86 valid object instances across 10 tabletop scenes. We compare against SAM3D, SAM3D+FoundationPose (SAM3D+FP), TRELLIS.2, and Hunyuan3D-2.1 under a shared calibrated camera view. Table~\ref{tab:object-reconstruction} reports per-instance metrics as mean $\pm$ standard deviation.  Color metrics are computed over the intersections of object masks, with RGB values normalized to $[0,1]$.

\begin{figure}[!t]
\centering
\includegraphics[width=0.95\columnwidth]{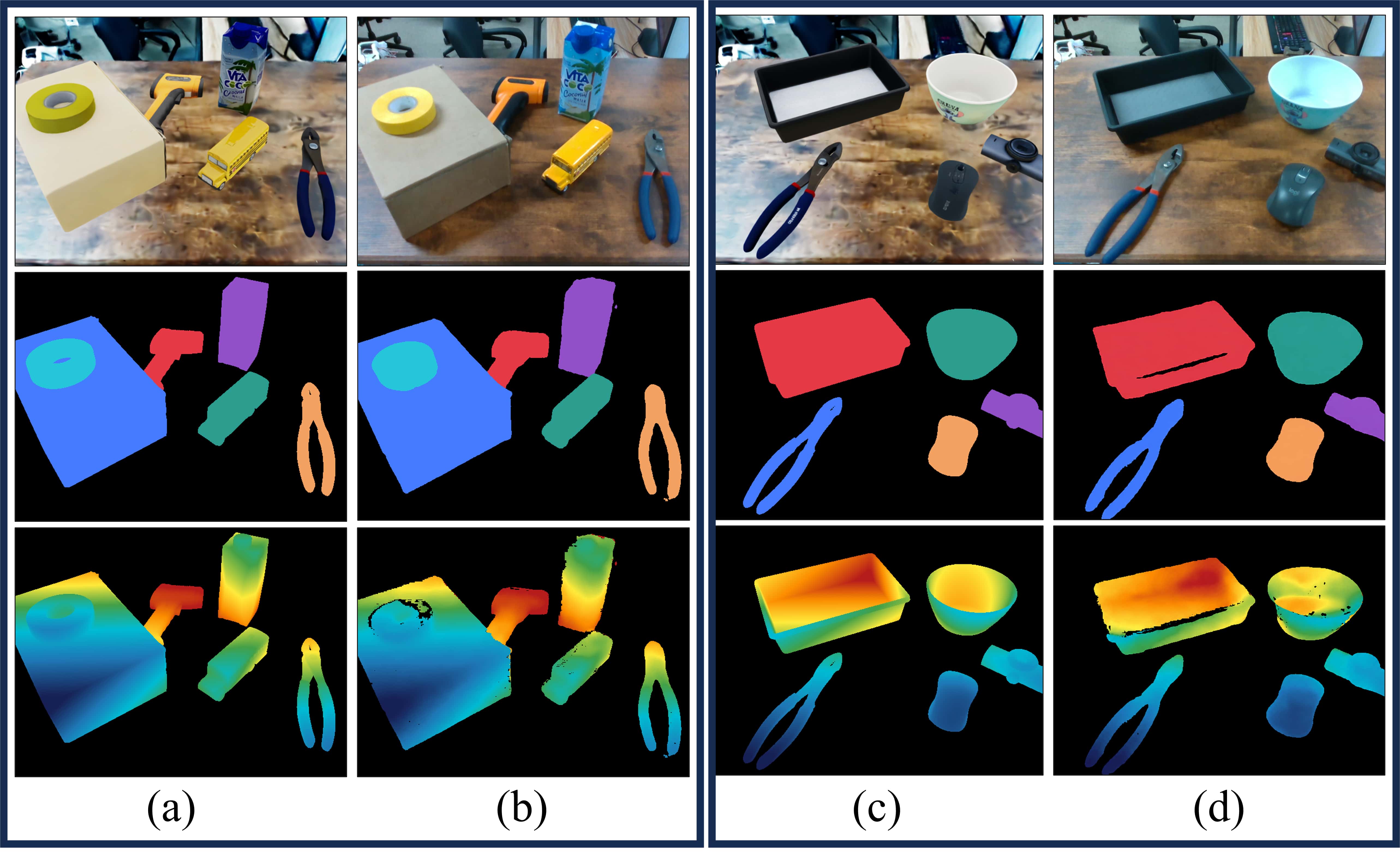}
\caption{Representative top-camera comparisons for two workspaces: (a, c) DEXTERA simulations and (b, d) corresponding real observations. Rows show RGB appearance, instance masks, and depth, from top to bottom.}
\label{fig:scene-reconstruction-comparison}
\end{figure}

\begin{figure}[!t]
\centering
\includegraphics[width=\columnwidth]{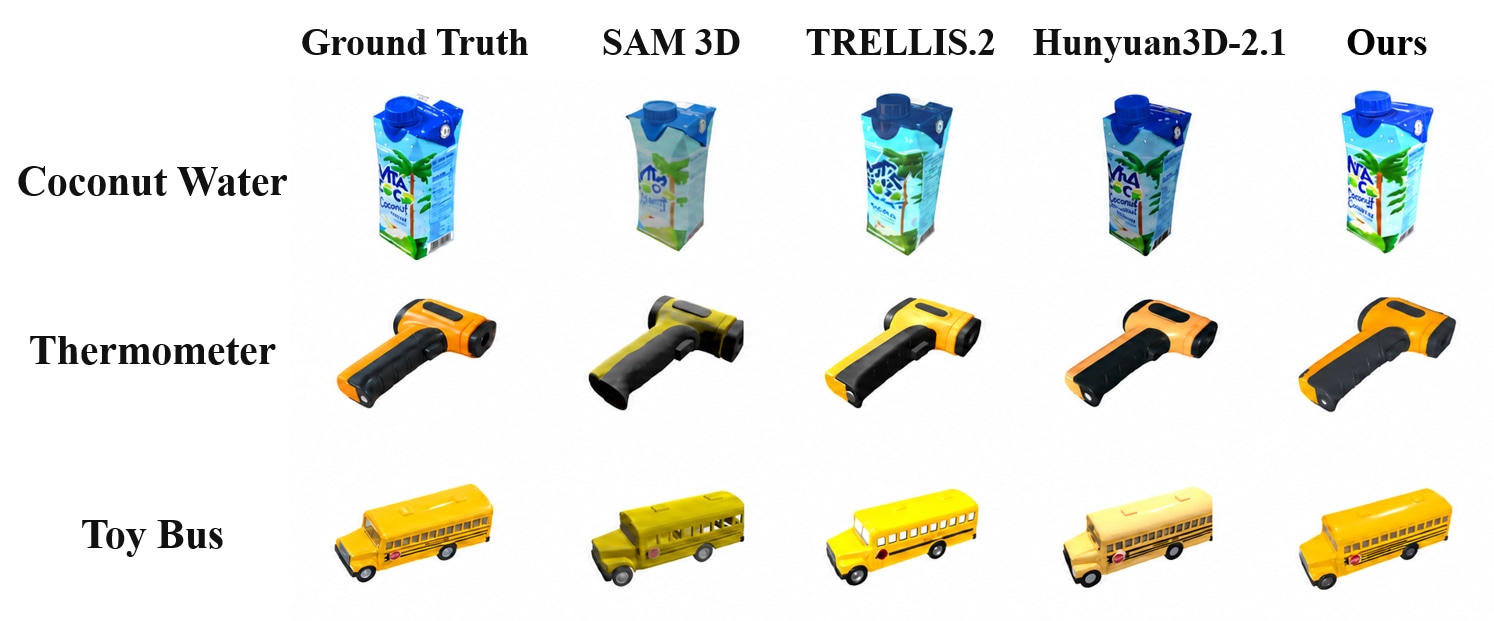}
\caption{Qualitative object-reconstruction comparison on a coconut-water carton, thermometer, and toy bus. Columns show the ground-truth object, SAM3D, TRELLIS.2, Hunyuan3D-2.1, and DEXTERA (ours). DEXTERA preserves object shape and appearance across geometrically and texturally distinct assets.}
\label{fig:mesh-reconstruction-comparison}
\end{figure}

DEXTERA achieves the best mean performance across all reported object metrics (Table~\ref{tab:object-reconstruction}). The improvements span image-plane alignment, RGB appearance, and 3D geometry. Combining FoundationPose with SAM3D (SAM3D+FP) improves 3D alignment but leaves appearance errors unresolved, highlighting the necessity of unifying generative completion and Sim(3) alignment. Qualitative scene layouts and mesh comparisons are shown in Fig.~\ref{fig:scene-reconstruction-comparison} and Fig.~\ref{fig:mesh-reconstruction-comparison}.

Background fidelity is evaluated separately outside object instance masks. As summarized in Table~\ref{tab:background-reconstruction}, the 3D Gaussian background yields high visual and spatial accuracy (PSNR $20.74$\,dB, Chamfer-L1 $6.33$\,mm), establishing a faithful background twin for static collision modeling.

\begin{figure}[!t]
\centering
\includegraphics[width=\columnwidth]{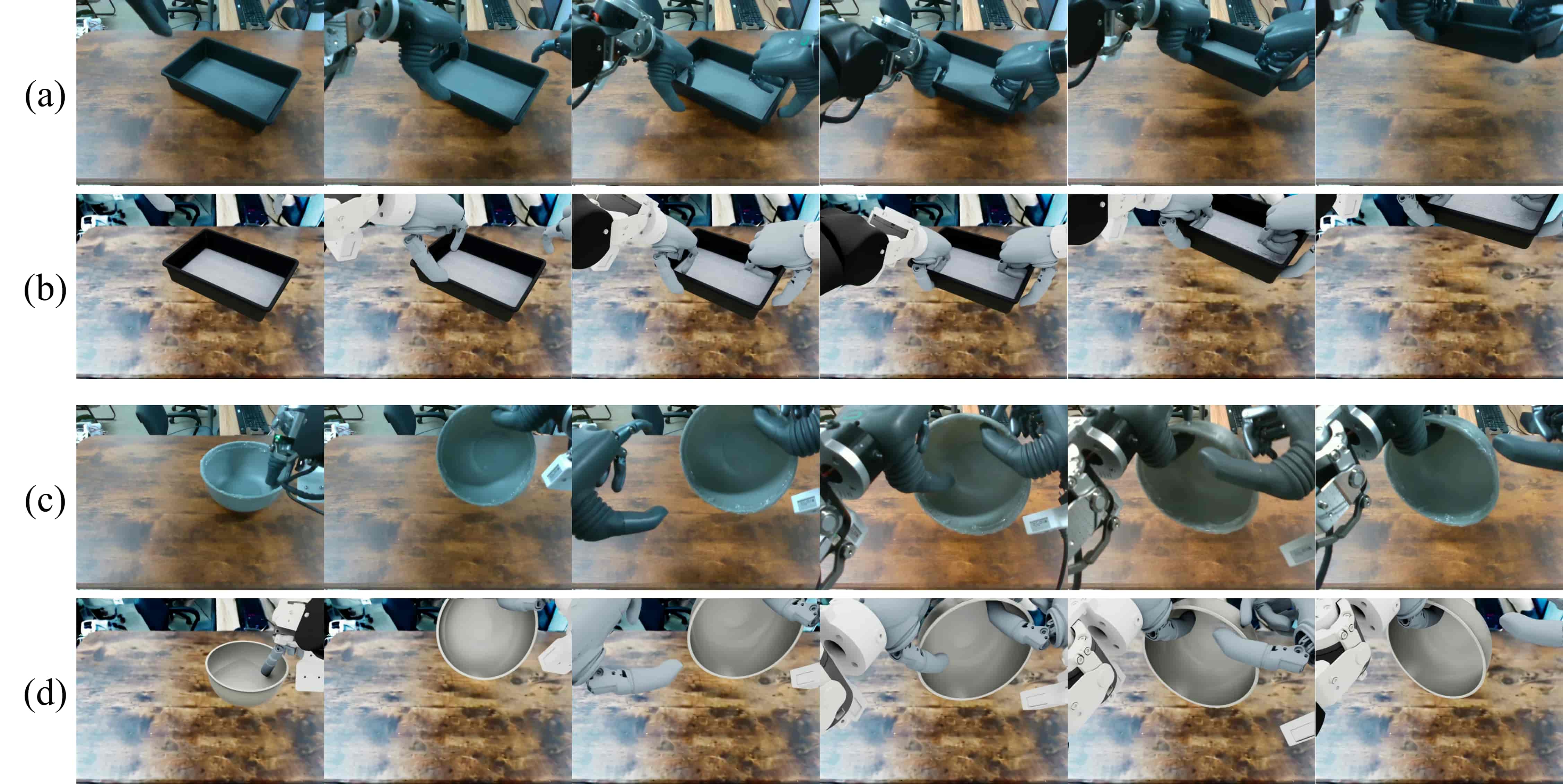}
\caption{Visual and interaction consistency during basket lifting and bowl handover. Rows (a, b) show simulated and real basket lifting, respectively; (c, d) show simulated and real bowl handover. Columns progress through successive interaction stages from left to right.}
\label{fig:realsim-replay}
\end{figure}

\begin{figure*}[!t]
\centering
\includegraphics[width=0.9\textwidth]{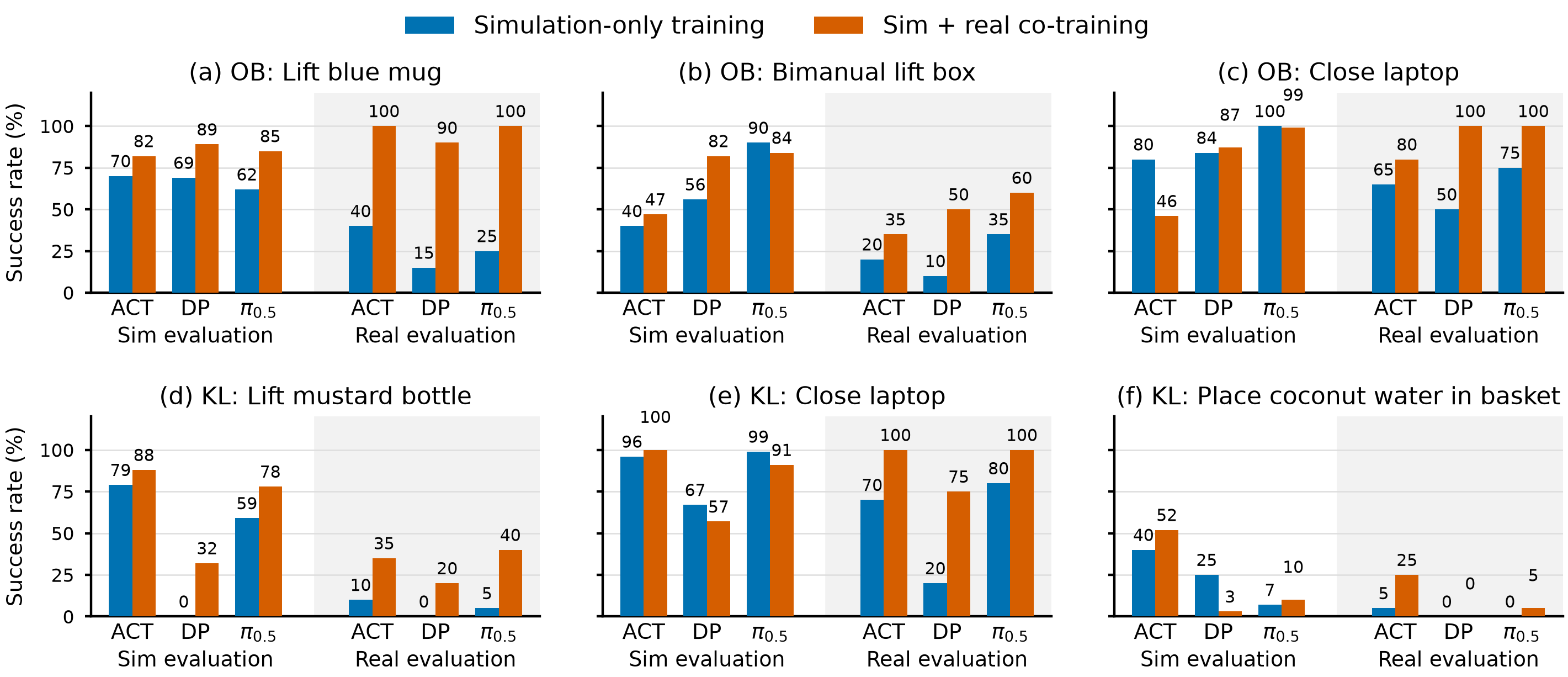}
\caption{Simulation-only training versus sim-real co-training for three policies on six task-embodiment pairs. Each panel separates simulation evaluation (left) from physical evaluation (right, shaded), with 100 and 20 trials per bar, respectively. Bar labels report success percentages. OB: OpenArm + BrainCo Revo1; KL: KUKA + LEAP Hand. Co-training improves physical success in 17 of 18 comparisons.}
\label{fig:training-data-comparison}
\end{figure*}

\subsection{Physical Consistency through Cross-Domain Replay}
\label{sec:exp-physics}
To assess whether simulated interactions preserve physical consistency when transferred to hardware, we perform cross-domain trajectory replays. At each physical reset, object tracking reconstructs the randomized real configuration in simulation, while object-centric trajectory synthesis adapts the demonstration to generate execution trajectories. For each task, we retain 30 simulation-successful rollouts and evaluate their execution on the physical robot.

Table~\ref{tab:physical-replay} reports an overall physical replay success rate of $87.95\%$ ($343/390$ trials) across contact-rich tasks (Fig.~\ref{fig:realsim-replay}), with four task configurations achieving $100\%$ success. Reorientation and bimanual box placement achieve lower success rates of $73.33\%$ each, potentially reflecting sensitivity to contact slippage and hardware compliance. These results validate that DEXTERA's reconstructed assets retain physical consistency for physical transfer.

\subsection{Policy Learning and Sim-to-Real Transfer}
\label{sec:exp-policy}
\paragraph{Setup}
We evaluate six policy architectures: ACT, Diffusion Policy (DP), BC-RNN, OpenVLA, $\pi_{0.5}$, and PPO Teacher--Student distillation (T--S). Ten VR demonstrations per task-embodiment pair are expanded to yield a total of 3,900 simulated and 390 physical trajectories across 13 task pairs. Imitation policies are trained via domain-balanced co-training with a 1:1 simulation-to-real mixture ratio and visual augmentations. Each evaluation condition comprises 100 simulation rollouts and 20 physical trials using unified action spaces and identical success criteria (Table~\ref{tab:policy-benchmark}).

\paragraph{Benchmark results}
Among all evaluated methods, $\pi_{0.5}$ achieves the highest mean physical success rate of 56.5\%, closely followed by ACT at 54.2\%. In contrast, OpenVLA averages 13.8\%, despite reaching 20\% on KUKA mustard-bottle lifting and 65\% on laptop closing. Performance varies substantially across tasks: while $\pi_{0.5}$ achieves 100\% physical success on OpenArm mug lifting and laptop closing, all methods top out at 15\% on toy-bus placement. Notably, high simulation success does not directly translate to real-world deployment. For instance, Teacher--Student distillation (T--S) achieves 96\% simulation success on OpenArm laptop closing, but drops to 30\% on hardware. Conversely, open-loop replay of simulation-successful rollouts yields 100\% (30/30) physical success on the exact same task (Section~\ref{sec:exp-physics}). This contrast highlights that executable simulated trajectories do not guarantee reliable closed-loop policy transfer.

Model capabilities also exhibit distinct task-specific strengths. On OpenArm bowl handover, ACT outperforms $\pi_{0.5}$ (65\% vs. 40\% physical success). Meanwhile, multi-step placement remains universally challenging across embodiments: no method exceeds 20\% on OpenArm bimanual box placement or 25\% on KUKA coconut-water placement. For the latter, poor performance even in simulation points to fundamental difficulties in learning the full grasp--transport--placement sequence, an error that compounds further on physical hardware. 

\paragraph{Simulation-only training and co-training}
Fig.~\ref{fig:training-data-comparison} evaluates ACT, DP, and $\pi_{0.5}$ across six selected task--embodiment pairs, highlighting zero-shot transfer capabilities and co-training effectiveness. Crucially, policies trained exclusively on simulation data achieve viable zero-shot sim-to-real transfer, attaining mean physical success rates of 35.0\% (ACT), 15.8\% (DP), and 36.7\% ($\pi_{0.5}$). Direct transfer proves particularly effective on structured tasks such as laptop closing, where $\pi_{0.5}$ achieves 75\% physical success on OpenArm and 80\% on KUKA. These results confirm that our reconstructed environments provide sufficient physical grounding to support zero-shot real-robot deployment.

Meanwhile, co-training with limited physical demonstrations significantly boosts real-world deployment performance, elevating physical success means to 62.5\% (ACT), 55.8\% (DP), and 67.5\% ($\pi_{0.5}$). Across all 18 policy--task comparisons, co-training improves physical success in 17 cases (with one tie), raising the overall physical mean from 29.2\% to 61.9\%. Beyond real-robot gains, co-training also partially enhances policy performance within the simulation environment, boosting the average simulation success rate from 62.4\% to 67.3\%. While task-specific trade-offs occasionally occur, such as ACT on OpenArm laptop closing where simulation success shifts from 80\% to 46\% as physical success rises from 65\% to 80\%, the overall evidence demonstrates that sim-real co-training substantially reinforces physical execution while preserving or enhancing simulation capability.

\section{CONCLUSION}
\label{sec:conclusion}
We presented DEXTERA, an automated real-to-sim-to-real framework for dexterous manipulation that connects single-image asset generation, metric scene--robot alignment, task primitive construction, trajectory synthesis, and policy deployment. Comprehensive evaluations across 13 task--embodiment pairs on two dexterous platforms demonstrate strong physical consistency, achieving a 343/390 (87.95\%) success rate in cross-domain trajectory replays on contact-rich interactions. Furthermore, simulation-only trained policies enable viable zero-shot real-world transfer, while simulation--real co-training elevates mean physical success from 29.2\% to 61.9\% across ACT, DP, and $\pi_{0.5}$ on six task pairs.

Despite these promising results, several limitations highlight important avenues for future work. First, asset generation is currently restricted to rigid and articulated objects, lacking support for deformable materials or fluid dynamics. Second, our evaluations focus on foundational task primitives, and the framework's efficacy on complex, long-horizon manipulation workflows remains to be validated. Third, persistent dynamics gaps and imprecise VLM-inferred physical parameters affect success rates on multi-step tasks. Fourth, although our framework supports room-scale background generation, current physical experiments are confined to stationary manipulation setups. Therefore, extending DEXTERA to mobile manipulation within larger workspaces remains a promising future direction. Finally, robot kinematics and joint parameters are currently assumed as known priors rather than inferred end-to-end. Future work will explore the potential of single-image robot morphology reconstruction as well.

\bibliographystyle{IEEEtran}
\bibliography{reference_brief}
\end{document}